\pdfoutput=1
\PassOptionsToPackage{table}{xcolor}
\documentclass[11pt]{article}

\usepackage{acl}

\usepackage{times}
\usepackage{latexsym}
\usepackage[T1]{fontenc}

\usepackage[utf8]{inputenc}

\usepackage{microtype}
\usepackage{graphicx}
\usepackage{multirow}
\usepackage[table]{xcolor}
\usepackage{pbox}
\title{The timing bottleneck: Why timing and overlap are mission-critical for conversational user interfaces, speech recognition and dialogue systems}

\author{Andreas Liesenfeld, Alianda Lopez, Mark Dingemanse \\
  Centre for Language Studies \\
  Radboud University, Nijmegen, The Netherlands \\
  \texttt{ \{andreas.liesenfeld,ada.lopez,mark.dingemanse\}@ru.nl} \\}
  
\begin{document}
\maketitle
\begin{abstract}
Speech recognition systems are a key intermediary in voice-driven human-computer interaction. Although speech recognition works well for pristine monologic audio, real-life use cases in open-ended interactive settings still present many challenges. We argue that timing is mission-critical for dialogue systems, and evaluate 5 major commercial ASR systems for their conversational and multilingual support. We find that word error rates for natural conversational data in 6 languages remain abysmal, and that overlap remains a key challenge (study 1). This impacts especially the recognition of conversational words (study 2), and in turn has dire consequences for downstream intent recognition (study 3). Our findings help to evaluate the current state of conversational ASR, contribute towards multidimensional error analysis and evaluation, and identify phenomena that need most attention on the way to build robust interactive speech technologies.
\end{abstract}

\section{Introduction}

Speech recognition (ASR) is a key technology in voice-driven human-computer interaction. Although by some measures speech-to-text systems approach human transcription performance for pristine audio \cite{stolcke_comparing_2017}, real-life use cases of ASR in open-ended interactive settings still present many challenges and opportunities \cite{addlesee_comprehensive_2020}. The most widely used metric for comparison is word error rate, whose main attraction —simplicity— is also its most important pitfall. Here we build on prior work calling for error analysis beyond WER \cite{mansfield_revisiting_2021,zayats_disfluencies_2019} and extend it by looking at multiple languages and considering aspects of timing, confidence, conversational words, and dialog acts. 

As voice-based interactive technologies increasingly become part of everyday life, weaknesses in speech-to-text systems are rapidly becoming a key bottleneck \cite{clark_state_2019}. While speech scientists have long pointed out challenges in diarization and recognition \cite{shriberg_errrr_2001,scharenborg_reaching_2007}, the current ubiquity of speech technology means new markets of users expecting to be able to rely on speech-to-text systems for conversational AI, and a new crop of commercial offerings claiming to offer exactly this. Here we put some of these systems to the test in a bid to contribute to richer forms of performance evaluation.

\begin{figure*}[!ht]
\centering
\includegraphics[width=\textwidth]{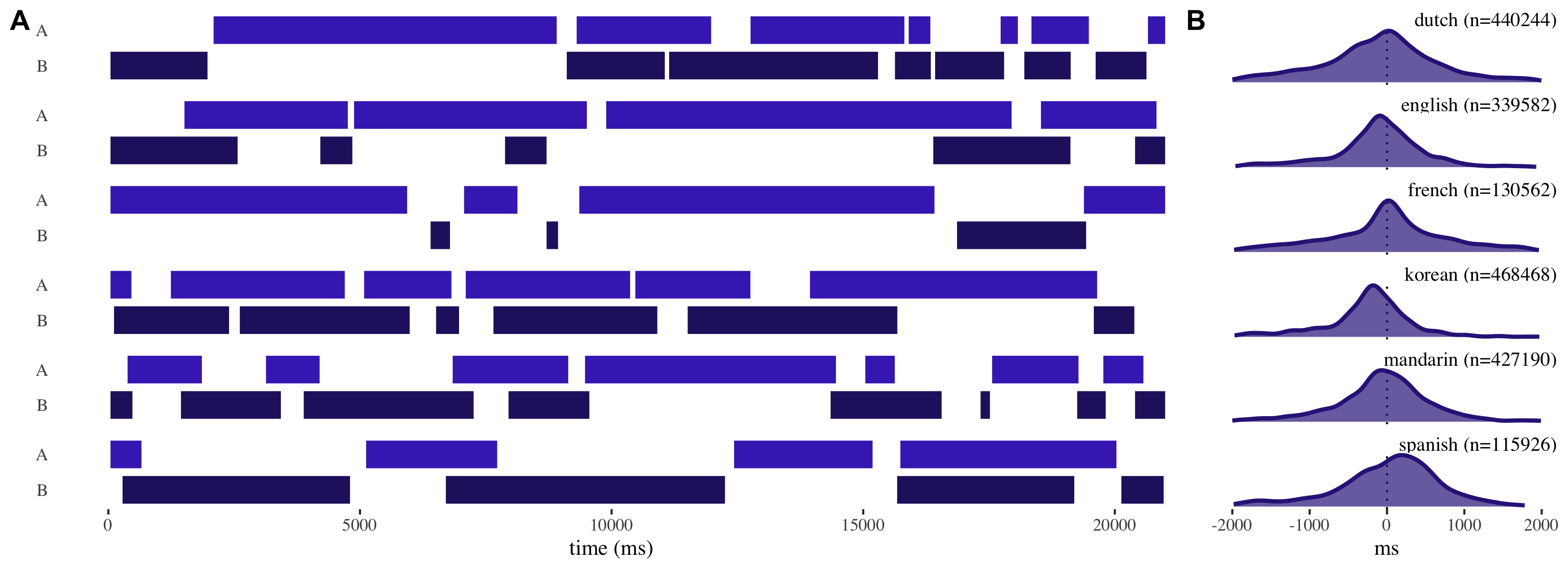}
\caption{\label{fig:human_timing} \textbf{A} Excerpts of 20 seconds of conversations in six languages, showing the short gaps and overlaps typical of human interaction. \textbf{B} Distribution of floor transfer offset times for the human-annotated test data across the same six languages, showing that the distributions are broadly normal and tend to peak around 0, with about as many turns occurring in slight overlap (negative values) as coming in after a slight gap (positive values).}
\end{figure*}

\subsection*{Related Work}

The struggles of achieving truly conversational speech technologies are well documented. Spontaneous, free-flowing conversations are effortless and efficient for humans but remain challenging for machines \cite{shriberg_spontaneous_2005, baumann_recognising_2017}. Speech-to-text systems face an interconnected set of challenges including at least voice activity detection, speaker diarization, word recognition, spelling and punctuation, code-switching, intent recognition, and more  \cite{suzuki_speech_2016,sell_diarization_2018, addlesee_comprehensive_2020,park_review_2022}. Each of these represents a choice point with downstream consequences that may be hard to predict. Perhaps this is why word error rate, despite its noted defects \cite{aksenova_how_2021, szymanski_wer_2020}, has gained the upper hand in ASR evaluation: it makes no assumptions and simply delivers a single number to be optimized. 

Speech scientists have long worked to supplement word error rate with more informative measures, including error analyses of overlap \cite{cetin_overlap_2006}, disfluencies \cite{goldwater_which_2010}, and conversational words \cite{zayats_disfluencies_2019, mansfield_revisiting_2021}.  This work has shown the importance of in-depth error analysis, and also brings home the multi-faceted challenges of truly interactive speech-to-text systems. As speech-to-text systems gain larger user bases, multilingual performance and evaluation becomes more important \cite{levow_developing_2021,blasi_systematic_2022,chan_training_2022,tadimeti_evaluation_2022}.  

The past decades of work on speech-to-text have led to remarkable improvements in many areas, and shared tasks have played an important role in catalyzing research efforts in diarization and recognition \cite{ryant_third_2021,barker_fifth_2018}. Still, we see opportunities for new contributions. Most work involves either non-interactive data or widely used meeting corpora, both of them quite distinct from the fluid conversational style people increasingly expect from interactive speech technology. When more conversational data is tested, it tends to be limited to English \cite{mansfield_revisiting_2021}, raising the question how large the performance gap is in a more diverse array of languages \cite{besacier_automatic_2014}. While most benchmarks still rely on word error rates, true progress requires more in-depth forms of error analysis \cite{szymanski_wer_2020} and especially a focus on the role of timing and overlap in speech recognition and intent ascription. Finally, the wide range of speech-to-text systems on offer in a time of need for robust conversational interfaces makes it important to know what current systems can and cannot do.


\section{Aims and scope}

A central question relevant at every moment of human interaction is \textit{why that now?} \cite{schegloff_opening_1973}, referring to the importance of position and composition in how people ascribe intent to communicative actions. For speech-to-text systems, in order to even approach this question, a key prerequisite is to detect \textit{who says what when}. This means that diarization, content recognition and precise timing are all highly consequential and best considered in tandem. 

Here we address this challenge by presenting a multipronged approach that lays some of the empirical groundwork for improving evaluation methods and measures. Using principles of black-box testing \cite{beizer_black-box_1995}, we evaluate major commercial ASR engines for their claimed conversational and multilingual capabilities. We do so by presenting case studies at three levels of analysis. Study 1 considers word error rates and treatment of overlaps. Study 2 looks into what goes missing and why. Study 3 looks into the repercussions for intent ascription and dialog state tracking. We show that across these areas, timing is both a mission-critical challenge and an ingredient for ways forward. 

\begin{figure*}
\centering
\includegraphics[width=\textwidth]{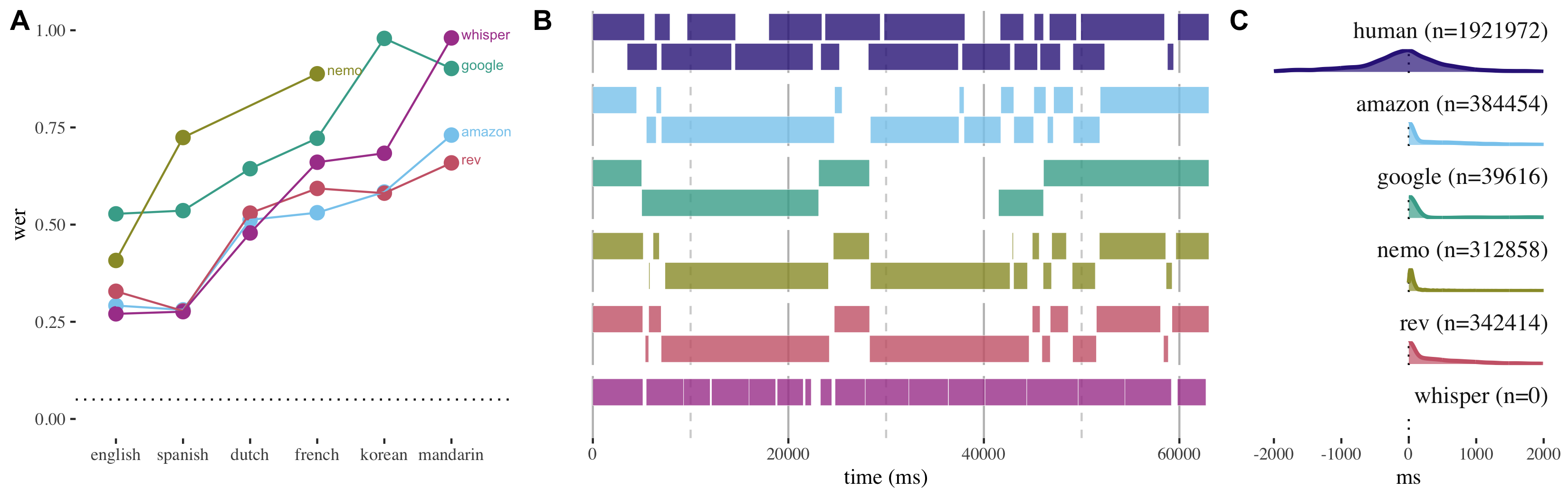}
\caption{\label{fig:study1_panel} \textbf{A} Word error rates (WER) for five speech-to-text systems in six languages. \textbf{B} One minute of English conversation as annotated by human transcribers (top) and by five speech-to-text systems, showing that while most do some diarization, all underestimate the number of transitions and none represent overlapping turns (Whisper offers no diarization). \textbf{C} Speaker transitions and distribution of floor transfer offset times (all languages), showing that even ASR systems that support diarization do not represent overlapping annotations in their output. }
\end{figure*}

\subsection*{Data and methods}

\textit{Data preparation.} We evaluate using a set of human-transcribed conversational data in multiple languages (Figure \ref{fig:human_timing} and Appendix A1). We take several steps to ensure the dataset makes for a useful evaluation standard: (1) we pick languages that all or most of the tested systems claim to support (English, Spanish, Dutch, French, Korean, and Mandarin); (2) we source conversational speech data from existing corpora with high quality human-transcribed annotations that were published as peer-reviewed resources; (3) we ensure audio files have comparable audio encoding and channel separation, (4) we curate human transcriptions and timing information of each dataset for completeness and accuracy, making sure that turn beginnings and ends are marked with at least decisecond precision (0.1ms); (5) we random-select one hour of dyadic conversations per language. More information on data sources and curation is available in this open data repository: \url{https://osf.io/hruva}. 

\textit{ASR system selection.} Following principles of black-box testing \cite{beizer_black-box_1995}, we test five widely used ASR systems, keeping data and testing methods constant to compare them to human transcription baselines. Functional testing does not require access to model code or training data, instead treating models as black boxes tested to specification \cite{ribeiro_beyond_2020}. Enabling independent verification and evaluation, it is a key method in the toolbox of NLP evaluation methods. 

We selected systems that claim to represent and handle conversational speech, and that offer multilingual support: (1) \href{https://aws.amazon.com/transcribe/faqs/}{Amazon Transcribe} 0.6.1, whose use cases include ``transcription of voice-based customer service calls" and ``generation of subtitles on audio/video content"; (2) \href{https://cloud.google.com/speech-to-text}{Google Cloud Speech-to-Text API}, using the \texttt{latest\_long} model meant for ``any kind of long form content such as media or spontaneous speech and conversations" (for French, Mandarin, and Spanish the long model is not available and we use the default model instead); (3) \href{https://nvidia.github.io/NeMo/}{NVIDIA NeMo} Quartznet15x15 for English and Conformer-CTC for French and Spanish, branded as a ``Conversational AI Toolkit'' that allows humans to ``interact naturally"; (4) \href{https://www.rev.ai/async}{Rev AI Asynchronous Speech-to-Text} API 2.17.1, which claims ``accurate speaker separation" and support for ``different speakers and conversations"; and (5) \href{https://github.com/openai/whisper}{Whisper}, a multilingual open-source neural net approaching ``human-level robustness and accuracy on English speech recognition". We collected the finest-grain data available for each of these systems, using \texttt{whisper-timestamped} \cite{louradour_whisper-timestamped_2023} to extract word-level timing from Whisper, and \texttt{pyannote.metrics} \cite{bredin_pyannotemetrics_2017} for speaker diarization with NeMo.

\subsection*{Study 1: WER and overlap in 6 languages}

\begin{figure*}[!ht]

\begin{minipage}{\textwidth}

\centering
\includegraphics[width=\textwidth]{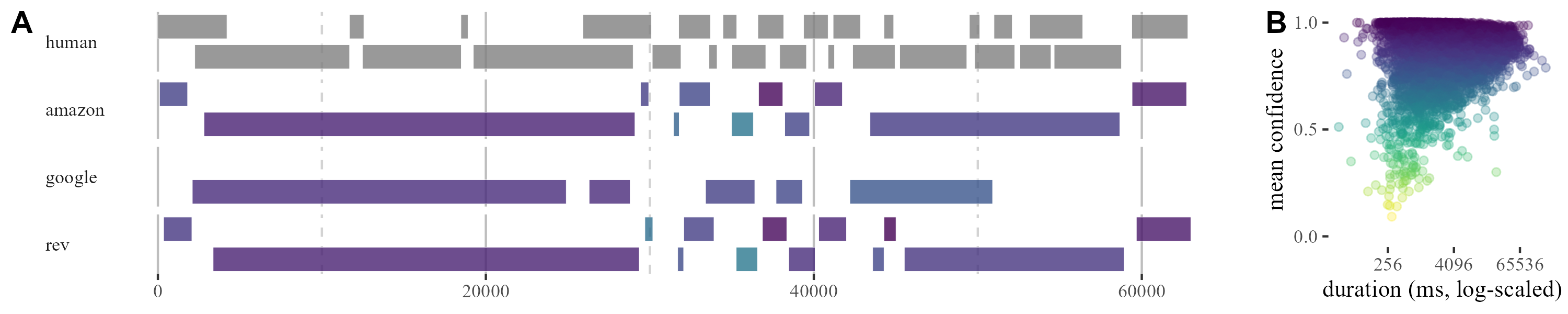}
\includegraphics[width=\textwidth]{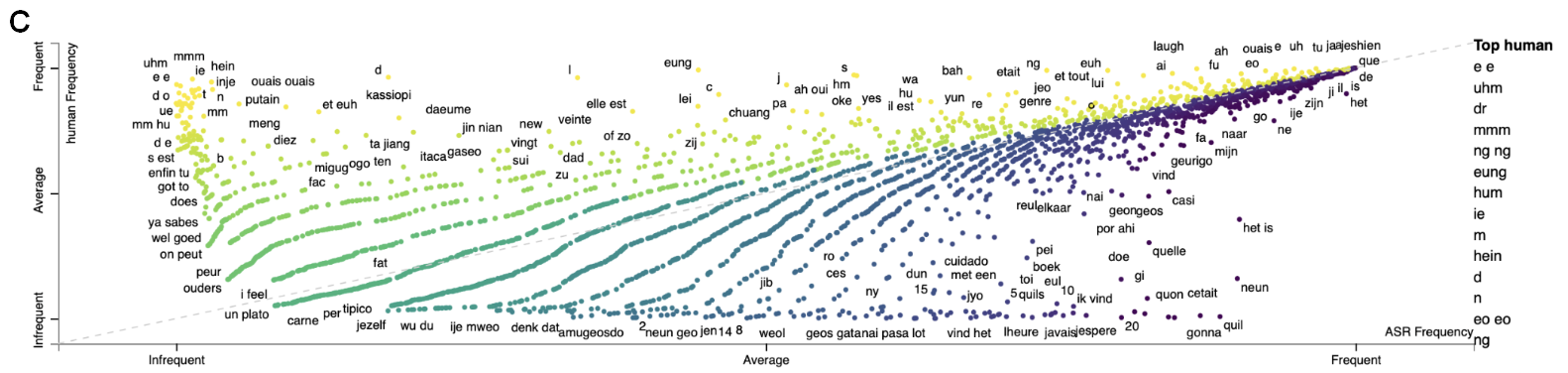}
\end{minipage}

\begin{minipage}{\textwidth}
\small
{\fontfamily{phv}\selectfont 
  \hspace{3pt} \textbf{D}}

\centering
\vspace{-5pt}
\begin{tabular}{l|lll}
 &
  Standalone Interjections &
  Function Words &
  Discourse Markers \\ \hline
\rowcolor[HTML]{E5E4E2} 
Dutch &
  {\color[HTML]{333333} \textit{uhm, hum, hu, uh ja, mm}} &
  {\color[HTML]{333333} \textit{'n, ie, d'r, da, 's}} &
  {\color[HTML]{333333} \textit{en uh, dat uh}} \\
English &
  {\color[HTML]{333333} \textit{mhm, uhhuh, hm, oh, wow}} &
  {\color[HTML]{333333} \textit{did, she's, that's, going to, he}} &
  {\color[HTML]{333333} \textit{yeah I, because}} \\
\rowcolor[HTML]{E5E4E2} 
French &
  {\color[HTML]{333333} \textit{hm hm, hein, ouais ouais, putain}} &
  {\color[HTML]{333333} \textit{c'était, qu'on, l', d', m'}} &
  {\color[HTML]{333333} \textit{euh tu,  et euh}} \\
Korean &
  {\color[HTML]{333333} \textit{ahyu, eung, ye, eo, jeogi}} &
  {\color[HTML]{333333} \textit{hae, gajigo, jeo, geuge, jal}} &
  {\color[HTML]{333333} \textit{geureonigga, geuraegajigo}} \\
\rowcolor[HTML]{E5E4E2} 
Mandarin &
  {\color[HTML]{333333} \textit{ng ng, ai, a, dui dui, er}} &
  {\color[HTML]{333333} \textit{la, wo wo, re, jiang, ya}} &
  {\color[HTML]{333333} \textit{shi er, gai, shi shuo}} \\
Spanish &
  {\color[HTML]{333333} \textit{eh, ah, he, claro, vale}} &
  {\color[HTML]{333333} \textit{o, eso, ahi, sea}} &
  {\color[HTML]{333333} \textit{o sea, sabes, verdad es}}
\end{tabular}%

\caption{\label{fig:study2_panel} \textbf{A} Sample minute of Korean conversation comparing human-transcribed and ASR annotations, the latter coloured by mean confidence rating. Shorter utterances and regions with more overlap are associated with lower confidence. \textbf{B} Mean confidence for ASR-transcribed utterances (n=17.563) by duration, showing that across all languages, low confidence scores are associated with shorter utterances. \textbf{C} Most characteristic elements in human-transcribed (yellow) and ASR transcribed (blue) conversational speech across all languages plotted by Scaled F-score, with the top most distinctive items for human transcripts on the right. \textbf{D} Top elements that are underrepresented or missing in ASR versus human-produced transcripts fall into three categories: short \textit{conversational interjections}, high frequency \textit{function words} (including contractions), and \textit{discourse makers}.}

\end{minipage}

\end{figure*}


\textit{Word error rates vary.} We find that word error rates for truly conversational speech vary widely but nowhere approach the oft-cited human baseline of 5\% transcription error (Figure \ref{fig:study1_panel}A, dotted line), a cross-linguistic replication of prior work on English \cite{mansfield_revisiting_2021}. Most speech-to-text systems have the lowest error rate for English, and even though all systems claimed multilingual support, all fare noticeably worse for typologically more different languages. 

\textit{Overlap is lost.} Human conversation typically features a rapid back-and-forth between participants, with a normal distribution of turn transition times centered around 0-200ms, and around half of all turns occurring in slight overlap (Figure \ref{fig:human_timing}; Figure \ref{fig:study1_panel}B-C, top). Tested ASR systems record substantially fewer speaker transitions and \textit{no} overlapping annotations. Distributions of speaker transition times show the consequences: current speech-to-text systems miss out on about half of the turns that occur in overlap. Descriptive statistics further corroborate this: by systematically not representing overlap, speech-to-text systems miss out on up to 15\% of all speech (or around 1 in 8 words), which results in an inaccurate picture of conversational content, structure, and flow (Table \ref{tab:table1} in Appendix).

\begin{figure*}[!ht]
\centering
\includegraphics[width=\textwidth]{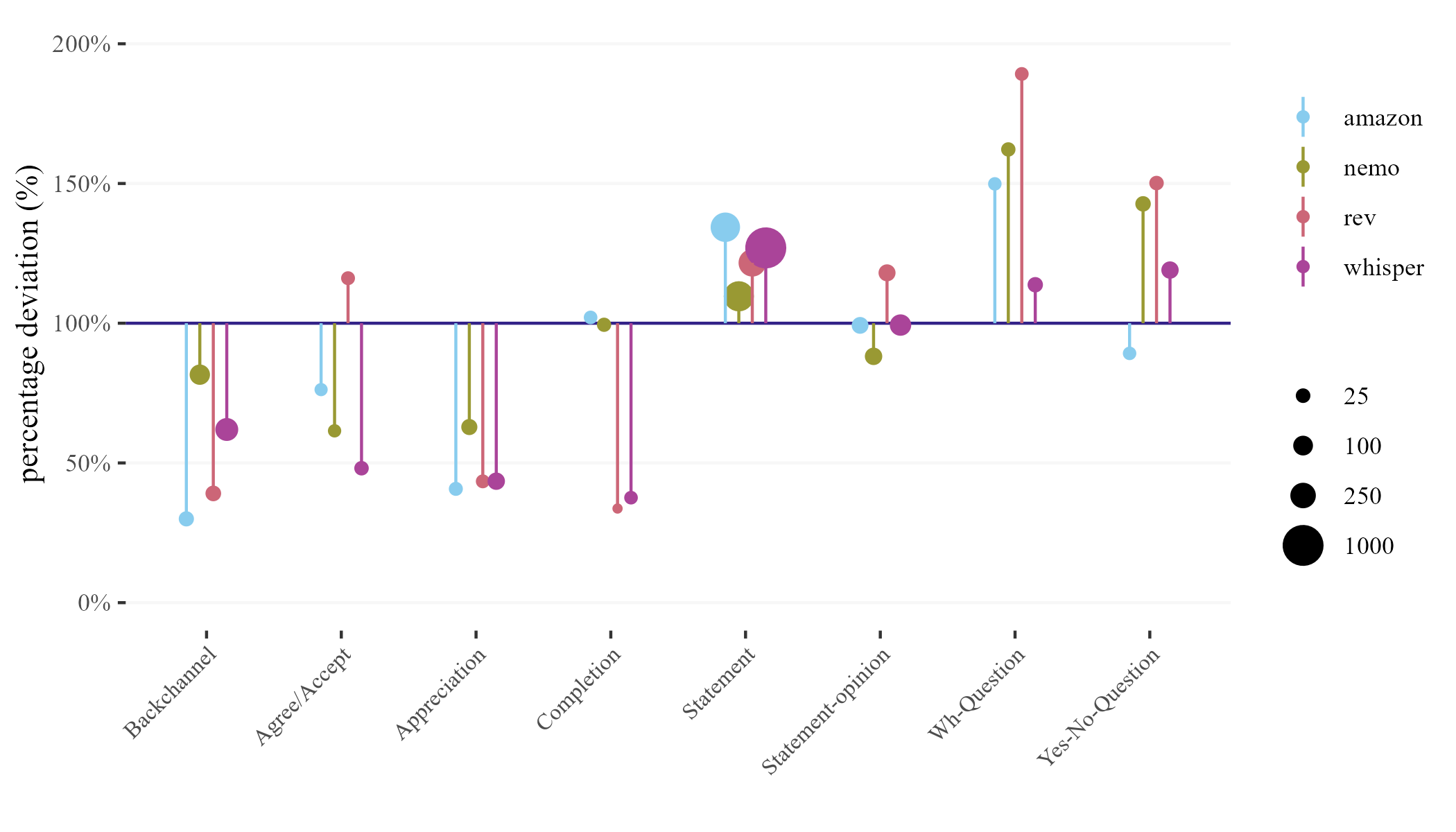}
\caption{\label{fig:dialog_acts} How different speech recognition engines warp dialog act classification in the same dataset of conversational English. For 8 frequent dialog acts, coloured lines show dialog acts based on ASR output deviate from those based on human transcripts of the same data (baseline). Dot size scales to number of times a tag is assigned. Only the most frequently assigned dialog acts (with at least 25 tokens in at least one dataset) are shown here. Mean absolute percentage deviations by ASR system: nemo 27.8\%, amazon 31.4\%, whisper 33.8\%, rev 47.4\%. }
\end{figure*}

\subsection*{Study 2: What goes missing and why}

\textit{Crosslinguistic replication.} Prior work on English has shown that it is especially short utterances and conversational words that go missing \cite{goldwater_which_2010,zayats_disfluencies_2019,mansfield_revisiting_2021}. Here we replicate this for all six languages in our sample (Figure \ref{fig:study2_panel}A). 

Confidence metrics supplied by three of the speech-to-text systems provide a novel view of this: regions with more overlap and shorter utterances often coincide, and both are associated with dips in word-level and utterance-averaged confidence scores (Figure \ref{fig:study2_panel}A-B). Across panels A, B and C, lighter coloured regions are associated with higher risk of being missed or misrecognized. 

\textit{Overlap-vulnerability and reduction.} In Figure \ref{fig:study2_panel}C, we compare human transcripts to ASR output using Scaled F-score \cite{kessler_scattertext_2017}, showing which items are underrepresented (top left) versus overrepresented (bottom right) in ASR output. We then take the top 15 most underrepresented items and inductively classify them as standalone interjections, function words, and discourse markers (Figure \ref{fig:study2_panel}D), following prior work \cite{zayats_disfluencies_2019,lopez_evaluation_2022}. We find that these categories provide good empirical coverage of what goes missing across all six languages in our sample.

Standalone interjections often occur in overlap-vulnerable contexts and are rare in ASR training data, often more formal and monologic \cite{liesenfeld_building_2022}. The category of function words mostly contains highly frequent bits of morphosyntax that may occur in overlap-vulnerable positions (as the Mandarin final particles \textit{la} and \textit{ya}) or that are likely to be phonetically reduced (as in Dutch and French contractions of pronominal forms). Finally, discourse markers are utterance-initial fragments that help direct the flow of a conversation. These too occur in overlap-vulnerable regions and are rare in ASR training data. 

\subsection*{Study 3: Consequences for dialog flow}

So far we have seen that the tested systems struggle with timing and overlap (study 1) and especially underrepresent conversational elements of speech (study 2). But how serious are the consequences for actual dialogue systems? One way of gauging this is to consider intent classification, a downstream task that is key to dialog state tracking and to virtually any practical application of voice UI  \cite{ye_structured_2022,gella_dialog_2022,jacqmin_you_2022}. 

As a minimal example, we use the Switchboard dialog act tagset \cite{stolcke_dialogue_2000} as implemented in the \texttt{dialogtag} Python library \cite{malik_dialogtag_2021} and apply it to (i) human transcripts and (ii) ASR transcripts of the same English subset of our data. By keeping the dialog tagger and the underlying data constant and  manipulating only the transcription method (human versus various ASRs) we make visible how reductions and variations introduced by speech recognition systems impact dialog act classification. We intentionally use the simplest possible dialog act tagger as a proof of concept. While several more sophisticated methods exist, every method is constrained by the data it can work with, and our goal here is to merely to make visible how ASR systems can impact intent ascription and dialog state tracking.

\begin{figure*}[!ht]
\centering
\includegraphics[width=\textwidth]{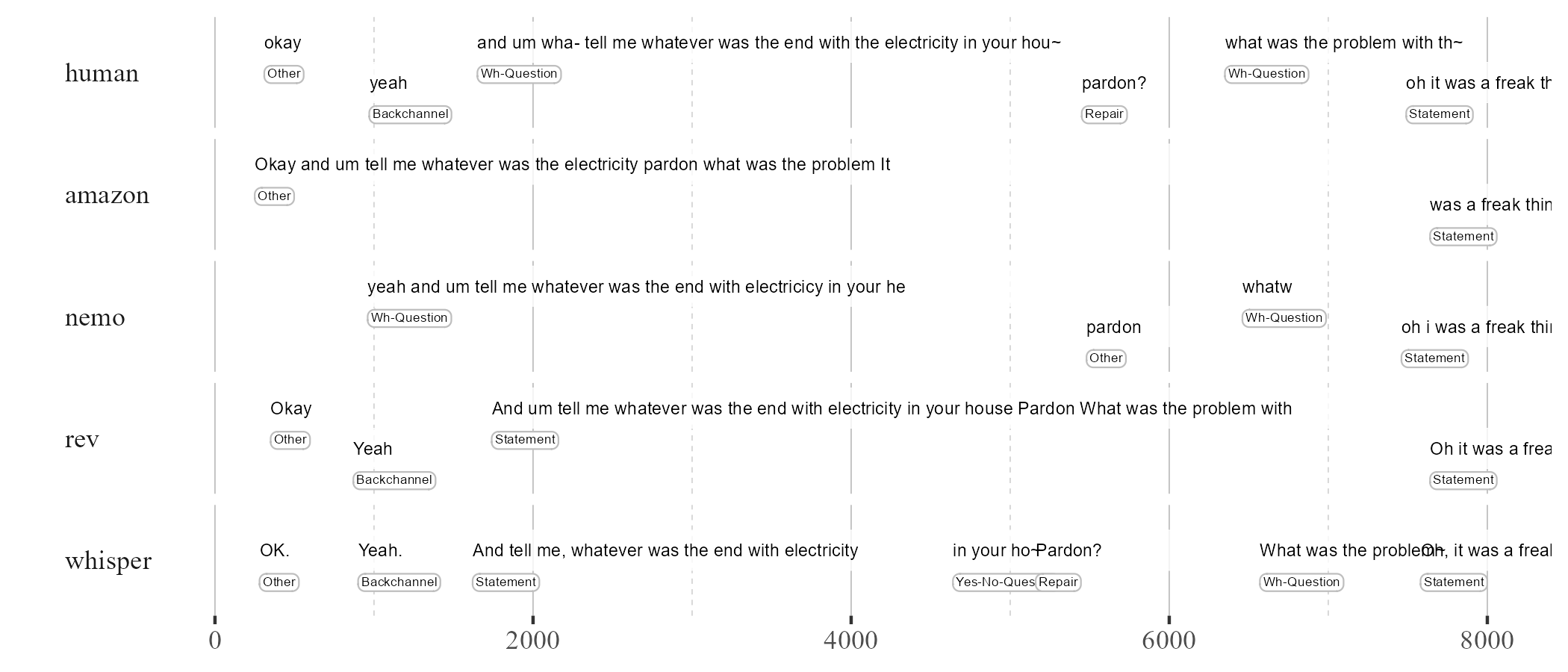}
\caption{\label{fig:dialog_acts_sample}Excerpt of 8 seconds of English conversation showing how differences in how speech-to-text systems carry out segmentation, diarization, and transcription have direct consequences for dialog act classification.}
\end{figure*}

We find that all ASRs warp dialog act classification outcomes in conversational English data (Figure \ref{fig:dialog_acts}). On average across the top 8 most frequently detected dialog act types, dialog act tags based on ASR output deviated between 27.8\% (nemo) to 47.4\% (rev) from tags based on human transcripts of the same data (this is absolute percentage deviation, i.e. including both overrepresentation and underrepresentation of dialog act tags). 

\textit{Interactionally consequential dialog acts.} Several highly interactionally relevant dialog act types are affected by speech-to-text systems. For instance (as expected based on Study 2), Backchannels and Agree/Accept tags are underrepresented across the board. This can be problematic for applications where it is important to keep track of user understanding and agreement during complex operations. Also, both the Wh-Question and Yes-No-Question dialog act tags tend to be overrepresented relative to the baseline. Since questions differ from other actions in the next moves they invite and expect, getting this wrong is directly consequential for any application in which user input is classified to determine relevant next actions.

\textit{What dialog act deformation looks like.} Figure \ref{fig:dialog_acts_sample} shows an excerpt of English conversation in its human-annotated version (top) and four ASRs, with dialog act annotations. We selected this excerpt because it illustrates many of the larger scale patterns of underrepresentation and overrepresentation evident in Figure \ref{fig:dialog_acts}.  Recall that dialog act tags are not supplied by the systems themselves, but applied to their output by \texttt{dialogtag}. Note that we speak of intent `ascription' rather than `recognition' to stress the fact that intents are often ambiguous and always provisional \cite{enfield_concept_2017}.

Starting with relatively short conversational elements, we find that \textit{yeah} is sometimes identified as a `Backchannel' (rev, whisper), sometimes merged with adjacent turns by the other speaker (nemo), and sometimes elided entirely (amazon) — the latter two cases exemplifying the reasons ASR output generally underrepresents this category. Similarly, \textit{pardon?} is variously identified as a `Repair' signal (whisper), sometimes missed as a separate action because it is merged into adjacent turns by the other speaker (amazon, rev), and sometimes tagged as `Other' (nemo), possibly because of punctuation. 

Moving on to more complex elements, we see that a lumping approach to segmentation can result in interactionally important dialog acts going undetected: Amazon merges two disparate turns, producing \textit{Okay and um tell me whatever was (...)}, which is tagged as Other. Meanwhile, a splitting approach, as Whisper appears to use, can lead to a fragment like \textit{in your house} being tagged as Yes-No-Question in whisper output, showing one likely cause of over-representation of such question tags. 

Disfluently produced questions can also pose problems: the utterance \textit{and um wha- tell me whatever was (...)}, which features a self-repaired fragment, is sanitized and identified as a Statement in its rev and whisper versions. In the nemo output, the same turn (though merged, as we saw above, with a preceding "yeah" by the other speaker) is correctly tagged as a Wh-Question.

Even in this simple proof-of-concept, we see that ASR output can affect the ascription and  classification of intents in various ways. This means that any real-world implementation relying on the systems tested here is hampered in its abilities to classify interactionally consequential social actions, making fluid interaction that much harder to achieve. Given the magnitude by which all tested ASRs deviate from human annotations in terms of timing, segmentation, diarization, overlap, and content, we expect similar kinds of distortion to appear in any systems for intent ascription and classification. 

\section{Discussion}
The ubiquity of voice interfaces coupled with reports of human parity in speech recognition might make robust voice-driven interaction seem within easy reach. Indeed, all major vendors now advertise speech-to-text pipelines that claim both multilingual ability and conversational utility. Here we put five such systems to the test and find that the results are bleak: word error rates are nowhere near the oft-claimed human parity; performance drops dramatically for languages other than English; precise timing and diarization is hard to come by; overlap is systematically ignored; conversational words go missing; and as a result, intent ascription and dialog state tracking are severely hampered. 

Commercial speech-to-text systems are frequently exposed to conversational settings, whether it is in home use, business meetings, or customer service interactions. Our results imply that these systems are likely to fall short of several of their intended applications. Word error rate does not sufficiently reflect the performance of speech-to-text systems in most real-life contexts. The erasure of conversational elements and inability to deal with overlap renders these systems effectively oblivious to important aspects of user feedback. Differences in diarization and turn allocation across systems also have strong effects on dialog act classification, with the  implication that switching vendors might have untold consequences for dialog state tracking and intent ascription. 

Our results show that current speech recognition systems privilege what is said over when it is said; and that even systems claiming conversational utility appear to treat the problem as fundamentally one of turning a rich tapestry of turns into running text. These text-first design choices become visible when exposed to the rapid turn-taking patterns of natural conversation — not only to analysts in case studies like this, but inevitably also to users, where they cause friction, interactional turbulence, and user dissatisfaction. The results are in line with recent arguments that the current language technology landscape is fundamentally built around monologic text instead of dialogical talk \cite{dingemanse_text_2022}. The rise of conversational interfaces motivates a course correction if not a refurbishing of the foundations. Here we hope to have shown that data from human interaction can inform such work. 

\subsection{Objections}

One might object that our test data is unreasonably tough, featuring open-domain informal conversation with rapid turn-taking and lots of overlap. We agree, but would counter that it is at the same time reasonably realistic: this is what typical human interactive behaviour look like. The brute facts of human interaction are something speech-to-text systems will need to reckon with if there is to be a chance of the ``natural interaction" and ``human-level robustness" promised by current solutions.

One might object that missing 1 in 8 words and having word error rates hovering around 50\% may not be fatal, depending on what goes missing. We agree, and point out that what goes missing here is crucial for interactive speech technology. Short recurring utterances like \textit{mmhm}, \textit{oh} and \textit{huh?} are the swiss army knife of conversational competence. These items enable robust communication and fluid coordination; to erase them is to rob users of their agency and to stunt the interactive capabilities of conversational technology.

One might object that dialog acts are an imperfect and language-specific way of looking at intent ascription, and that automated tagging based on form alone does not do justice to the situatedness of action \cite{rollet_talk_2020,parret_essential_1981}. We agree, and have picked dialog acts merely as a proof-of-concept to illustrate the more general problem of garbage in, garbage out: defective diarization, missing words, and neglect of timing will hamper any form-based methods for intent ascription and dialog state tracking.

\subsection{Limitations}
We are aware of the following limitations.

First, the human reference data is internally quite diverse, differing in recording setting and audio quality. This makes comparisons across datasets harder, so we have refrained from drawing strong comparative conclusions about possible differences across corpora and languages, instead focusing on recurring patterns of what goes missing and why. 

Second, we have not collected baseline measures for non-conversational data, making it hard to estimate how large the performance offset really is relative to more typical word error rate studies. Doing this would require a parallel data collection and curation exercise for each of the languages included in our study, which is outside our scope here but represents a good target for future work.

Third, given our choice to evaluate commercial vendor pipelines, we are unable to examine or report details about ASR system architectures, model parameters, and confidence score calculations. This is a necessary consequence of black-box testing. While direct access and manipulability offer important advantages from an engineering perspective, we nonetheless think it is also important to document and evaluate the performance of widely used commercial solutions. 

Fourth, we have only considered the timing information provided in ASR results, not the latency at which the results themselves are delivered. The latency of ASR systems at runtime imposes another formidable bottleneck on voice-driven conversational interfaces, especially as long as they use end-pointing methods, where response planning only starts when an utterance end is detected with some probability. User-perceived latency is the single biggest determinant of people's satisfaction with voice assistants \cite{shangguan_dissecting_2021,bijwadia_unified_2023}. Collecting realistic latency data would require implementing the tested systems in a voice UX environments with human users, which is beyond the scope of this paper (but see \citet{aylett_pilot_2023}). Empirical work on dyadic and multi-party interaction can show how people realize low latencies in real time. This is a high bar to meet, and it likely requires a radical overhaul of ASR systems towards incremental processing \cite{skantze_turn-taking_2021}.

\subsection{Recommendations}
The interconnectedness of all relevant processes in speech-to-text systems means that any quick fix likely has adverse consequences elsewhere. For instance, it is possible to improve diarization error rates by detecting and removing all overlap  \cite{boakye_twos_2008} — but this means throwing out at least 15\% of the data (as we show), putting human parity out of reach. Likewise, one may seek to reduce word error rates and interactional turbulence by excluding interjections \cite{papadopoulos_korfiatis_primock57_2022}, but this comes at the cost of losing all opportunity of rapid real-time user feedback. Our recommendations therefore focus on broadening the empirical basis, overcoming siloization, doing more in-depth evaluation, and incrementalizing architectures. 

\textit{Improve ecological grounding.} The most widely used datasets for training ASR systems still consist mostly of monologic read speech in well-resourced languages. For ASR systems to gain headway in truly interactive settings, they need to be exposed to more data that is closer to everyday language use in terms of linguistic diversity, conversational style, and participation \cite{aylett_you_2023}. Fortunately, such data is available for an ever-wider range of languages \cite{liesenfeld_building_2022}.

\textit{Overcome siloization.} In a field as large and varied as automatic speech recognition, some degree of specialization is inevitable, but true progress requires working together across disciplines. As we have shown here, engineering choices in voice activity detection directly affect dialog flow, and conversation designers benefit from knowing the limitations of word error rates and the importance of overlap. Reducing the siloing of knowledge will be crucial for resolving theoretical and practical challenges of speech recognition in the era of conversational interfaces.

\textit{Value qualitative error analysis}. Simple metrics make for attractive optimisation goals, but are always vulnerable to mindless metrics gaming: when a measure becomes a target, it ceases to be a good measure \cite{strathern_improvement_1996}. Qualitative error analysis and thorough human evaluation will remain important to truly get a handle on what goes wrong and how things can be improved \cite{szymanski_wer_2020}. This means incentives must be shifted to reward meaningful forms of evaluation over SOTA-chasing \cite{rogers_changing_2021,church_emerging_2022}. It also means there is room for more exploratory methods, such as the dialog act classification measure we have begun to explore here.

\textit{Develop multidimensional evaluation.} The downsides of word error rates have led to a flowering of alternative measures \cite{errattahi_automatic_2018, bredin_pyannotemetrics_2017}. In time, the field will benefit from a degree of consolidation, and holistic evaluation of speech-to-text systems will require taking into account a wider range of measures, including but not limited to diarization, timing, duration, overlap, coverage, phonology, spelling, and word error rate. Empirical and modelling work is needed to arrive at composite evaluation measures that are precise, reproducible and meaningful.

\textit{Strengthen incremental approaches.} Even if diarization quality, overlap detection and word error rates would come closer to human performance, the runtime latency of speech recognition stands in the way of fluid interactivity. To approach the rapid turn-taking and functional overlap that makes human interaction so flexible, voice-driven user interfaces will likely have to be designed as incremental architectures \cite{schlangen_general_2011}. Promising work in this domain exists \cite{baumann_recognising_2017,addlesee_comprehensive_2020,addlesee_understanding_2023}, and this represents an important growth area. 

\textit{Use timing when available}. Current systems at least provide timing for non-overlapping stretches of talk, but even that is rarely used for intent ascription. This despite the fact that we know timing alone can change the interpretation of a turn like ``Sure.", with longer delays flipping its polarity from positive to negative \cite{roberts_identifying_2013}. Building on insights like this, timing might be used to improve at least some elements of intent ascription. Likewise, known facts about relative durations of turns and silences could be used to make empirically informed decisions about when to lump versus split speech material in ASR output.

\section{Conclusion}
When you're a voice-driven conversational agent, life comes at you fast, and talk comes at you faster. We have presented evidence and arguments to support our contention that timing is more than a nice-to-have for any truly conversational system: it is mission critical and despite decades of attention from speech scientists remains largely unsolved today. But rather than despair we take our findings as an opportunity to identify areas where novel work can make big differences. While diarization remains hard in real-life settings, representing overlap instead of erasing it is likely to offer meaningful improvements. While overlap-vulnerable elements will always remain acoustically at risk, exposing ASRs to more ecologically valid training data and abandoning text-based sanitizing techniques will likely improve the recognition of short conversational elements. And while intent ascription will always be hampered by missing data, taking timing into account will enable new gains.

Dealing with conversational words computationally is hard: not only are their forms short and prone to overlap, their meanings are cognitively demanding and interactionally subtle. A focus on information and sentence structure over interaction and sequential organization has long enabled us to look away from these elements. As conversational words are backgrounded as `backchannels' and the artful interweaving of turns is classified as mere `overlap' if not `noise', it becomes easy to lose sight of the intricacies of human interaction. One way to see this paper is as contributing to a reframing that is underway in the language sciences at large: a reframing that foregrounds talk over text, that attends to interaction alongside information, and that recognizes the key role of timing. Timing is the secret sauce that can turn text into talk, chat into conversation, and perhaps, one day, clunky bots into fluid interactive tools.


\section*{Acknowledgements}
This work was funded by Dutch Research Council talent grant 016.vidi.185.205 to MD. We thank four anonymous reviewers for helpful comments, and JP de Ruiter for a discussion of the downsides of dialog acts.

\bibliography{elpaco_shared}
\bibliographystyle{acl_natbib}

\onecolumn
\appendix
\section{Appendix}
\label{sec:appendix}

\subsection{Datasheets}
Table \ref{tab:datasheets} shows the different corpora used in the study, detailing how many conversations were included and their total lengths in minutes. Every language contains approximately one hour worth of conversations, and when feasible, different interactional settings were incorporated (resulting to two corpora for Dutch). Each processing step is reflected in the processing pipeline avaliable in the repository, which also includes a datasheet \cite{gebru_datasheets_2021} and instructions on how to replicate the study given access to the data. For Dutch and Spanish, the evaluation datasets are freely available for academic research purposes. For English, French, Korean and Mandarin, the study repository provides information how to obtain the datasets used: \url{https://osf.io/hruva}. 

\begin{table}[ht!]
\centering
\resizebox{\textwidth}{!}{%
\begin{tabular}{l|lll}
Language & Corpus                                                                       & Conversations (n) & Length (mins) \\ \hline
\rowcolor[HTML]{E5E4E2} 
\cellcolor[HTML]{E5E4E2}                        & The Corpus of Spoken Dutch (CGN) \cite{taalunieCorpusGesprokenNederlands2014} & 3 & 30.11 \\
\rowcolor[HTML]{E5E4E2} 
\multirow{-2}{*}{\cellcolor[HTML]{E5E4E2}Dutch} & IFADV Corpus \cite{vansonIFADVCorpusFree2008}                                 & 2 & 29.97 \\
English  & CALLHOME American English \cite{canavanalexandraCALLHOMEAmericanEnglish1997} & 6                 & 60.25         \\
\rowcolor[HTML]{E5E4E2} 
French   & Nijmegen Corpus of Casual French \cite{torreiraNijmegenCorpusCasual2010}     & 6                 & 60.39         \\
Korean   & CALLFRIEND Korean \cite{canavanalexandraCALLFRIENDKorean1996}                & 4                 & 59.99         \\
\rowcolor[HTML]{E5E4E2} 
Mandarin & CALLHOME Mandarin Chinese \cite{canavanalexandraCALLHOMEMandarinChinese1996} & 6                 & 60.20         \\
Spanish  & Glissando Corpus \cite{garridoGlissandoCorpusMultidisciplinary2013}          & 6                 & 60.34        
\end{tabular}%
}
\caption{Corpora used in the study, with each language represented by approximately one hour of informal conversations.}
\label{tab:datasheets}
\end{table}

\subsection{Study 1 methods}
For both the human and ASR-transcribed data we calculate turn transition times in ms, number of speaker transitions, and the presence and duration of overlaps. For error analysis at the content level, we removed punctuation and excluded tags for non-speech behavior such as [laugh] and [breath] to bring all transcripts to a more comparable format. We used \texttt{\href{https://pypi.org/project/cleantext/}{cleantext}} for pre-processing and \texttt{whitespace} for tokenizing. We then calculated word error rate (WER) using \texttt{\href{https://github.com/jitsi/jiwer}{jiwer}}, and for a more in-depth investigation on the differences between human and speech-to-text annotations, we adopt Scaled F-score \cite{kessler_scattertext_2017} as a metric of n-gram salience scoring.

\subsection{Study 1 detailed results}
Table \ref{tab:table1} provides a more detailed look at key differences between human transcriptions and ASR output across the six languages in our sample. For every language, it lists the mean amount of speech covered by the transcriptions (coverage); the mean total number of words in the transcripts (words); the mean turn duration in milliseconds; and the mean percentage of overlapping annotations.

\begin{table*}[!hb]
\centering
\begin{tabular}{l|lllll}
 \pbox{20cm}{Human vs ASR }
 & \pbox{20cm}{Coverage (min)}    
 & \pbox{20cm}{Words (n)}                     
 &  \pbox{20cm}{Turn duration  (ms)}     
 & \pbox{20cm}{Overlap (speech \%)}                      \\ \hline
\rowcolor[HTML]{E5E4E2} 
Dutch   
&  63
&  12023 
&  2840 
& 13.4 \\ 
\rowcolor[HTML]{E5E4E2}                   
& 47                       
& 9396 
& 5897
& 0 \\

 English   
&  65
&  13895 
&  2811 
&  12.6 \\ 
                
& 55                      
& 10994   
& 6647   
& 0 \\ 

\rowcolor[HTML]{E5E4E2} 
 French  
&  64
&  13564 
&  4357 
&  14.4 \\ 
\rowcolor[HTML]{E5E4E2}              
& 49                     
& 8359   
& 7042   
& 0 \\

 Korean   
&  74
&  9632 
&  3280 
&  20.8 \\ 

& 43                      
& 5923 
& 4186 
& 0 \\ 

\rowcolor[HTML]{E5E4E2} 
 Mandarin  
&  66
&  15349 
&  2538 
&  15.8 \\ 
\rowcolor[HTML]{E5E4E2}             
& 53                        
& 8188 
& 7301   
& 0 \\ 

 Spanish   
&  63 
&  11868 
&  4620 
&  10.5 \\ 
                  
& 57                      
& 10177 
& 7534 
& 0 \\ 
\end{tabular}
\caption {\label{tab:table1} Comparison of human (top) and ASR transcripts (bottom) in each language in terms of coverage (amount of speech transcribed (in minutes), number of words, mean duration of each conversational turn (ms), and percentage of overlapped annotations relative to the length of the whole conversation. Human annotations add up to 395 minutes of transcribed speech; ASR-produced annotations for the same data on average add up to only 304, or 77\% of the observed speech. }
\end{table*}

\end{document}